\documentclass{article}
\usepackage{ital-ia}
\usepackage[utf8]{inputenc}
\usepackage[T1]{fontenc}

\usepackage{times}
\usepackage{soul}
\usepackage{url}
\usepackage[hidelinks]{hyperref}
\usepackage[utf8]{inputenc}
\usepackage[small]{caption}
\usepackage{graphicx}
\usepackage{amsmath}
\usepackage{booktabs}
\usepackage{algorithm}
\usepackage{algorithmic}
\title{AI-based Data Preparation and Data Analytics in Healthcare: The Case of Diabetes}

\author{
    Marianna Maranghi\and
    Aris Anagnostopoulos\and
    Irene Cannistraci\and
    Ioannis Chatzigiannakis\and
    Federico Croce\and
    Giulia Di Teodoro\and
    Michele Gentile\and
    Giorgio Grani\and
    Maurizio Lenzerini\and
    Stefano Leonardi\and
    Andrea Mastropietro\and
    Laura Palagi\and
    Massimiliano Pappa\and
    Riccardo Rosati\and
    Riccardo Valentini\And
    Paola Velardi\\
    \affiliations
    Sapienza Information-Based Technology InnovaTion Center for Health (STITCH)
    \\Sapienza University of Rome
    \emails
    \emph{name.surname}@uniroma1.it
}

\begin{document}

\maketitle

\begin{abstract}
    The Associazione Medici Diabetologi (AMD) collects and manages one of the largest worldwide-available collections of diabetic patient records, also known as the AMD database. This paper presents the initial results of an ongoing project whose focus is the application of Artificial Intelligence and Machine Learning techniques for conceptualizing, cleaning, and analyzing such an important and valuable dataset, with the goal of providing predictive insights to better support diabetologists in their diagnostic and therapeutic choices.
\end{abstract}

\section{Introduction}
In this project, we propose to apply data mining and machine learning methodologies to diabetes, one of the most common chronic diseases affecting hundreds of millions of people worldwide. The study will take advantage of one of the largest worldwide-available collections of diabetic patient records (the AMD database). It was recently made available by the Associazione Medici Diabetologi (AMD) and AMD Foundation to the Sapienza information-based Technology InnovaTion Center for Health (STITCH) at Sapienza University of Rome. Since 2005 AMD has promoted the creation of a network of diabetes outpatient clinics using the same software for the electronic medical record (EMR) collection. While this application was used for the clinical management of patients, it had the final aim to promote and improve the quality of care of diabetic subjects. Periodic data extraction from the EMR has been used for monitoring the quality of care indicators. This activity has produced, over the years, a systematic improvement of all the indicators considered and has proved to be cost-effective~\cite{Giorda2014}. Today the network comprises about 320 diabetes centers, and results are published every year in the AMD annals. 
However, the Annals database also represents a valuable source of observational research data in order to deepen many critical aspects of this chronic disease, such as the identification of groups of patients with different clinical courses, the short term prediction of the progression of the disease, and testing the effectiveness of innovative drugs in real-world practice.

Recently, machine learning techniques have been used to derive new insights, prevent adverse outcomes and support medical decision-making.  
This project is driven by the awareness that state-of-the-art machine learning techniques may provide predictive insights to better support clinicians in their diagnostic and therapeutic choices. In fact, as the science of diabetes advances, big data and machine learning represent a tremendous opportunity in diabetes care and prevention. However, the main limitations of these projects are the relatively small number of patients for which data were available (hundreds or thousands record) and the limited time interval during which data are collected. This limitation also holds for patient health records on several diseases~\cite{Johnson2016,Pollard2018}.In this project, we have the opportunity to overcome the above limitations since the available database contains data of about 600,000 patients collected over a time interval of 15 years (2005-2020).  
We believe that the availability of the AMD database represents a unique opportunity for an interdisciplinary team of scientists in diabetes, data analytics, and machine learning to advance state of the art in the study of this important chronic disease.

\section{Data Preparation}
The data shared by AMD is of inestimable value because it represents a very large amount of real data coming from real medical examinations on patients affected by diabetes. Nevertheless, due to the intrinsic heterogeneity of this data, it was not trivial to make it consistent and exploitable for meaningful data analytics. Indeed, the data were structured according to specific application needs and were not necessarily suitable for and were somewhat difficult to interpret for non-experts of the domain. For this reason, the two initial goals of our work were: (i) making a shared common knowledge of the information asset explicit; and (ii) building a central gathering point for future data analytics tasks. The actions we decided to take for reaching these goals are manifold and were conducted before any other kind of analysis, so that everyone had a common interpretation of the data, and worked on a cleaner and more consistent version of it.

\subsection{Domain ontology and data modeling} 
\begin{figure}[h]
    \centering
    \includegraphics[trim=5cm 0 2cm 0 clip, page = 1, scale=0.32, angle=90]{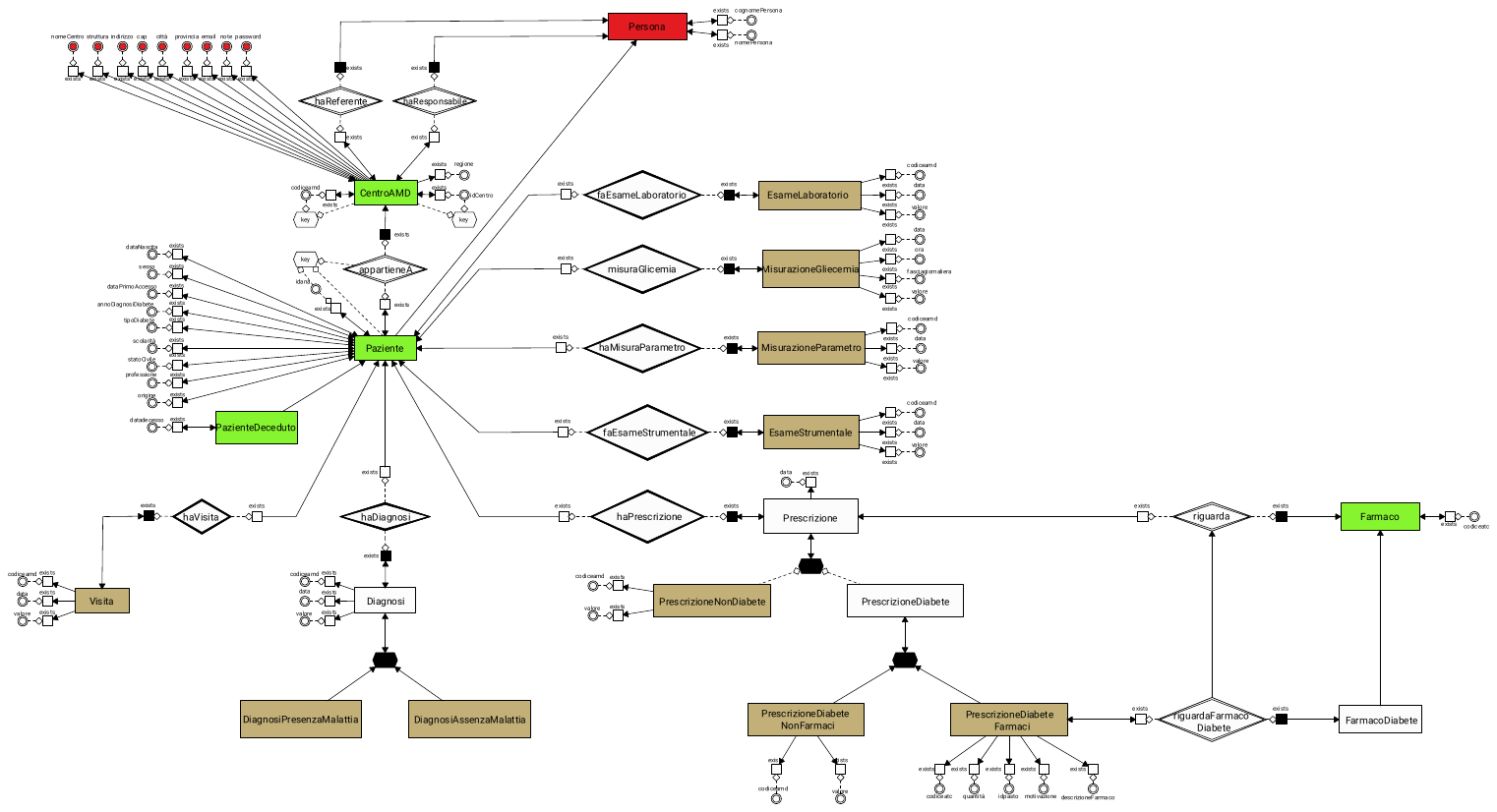}
    \caption{The ontology representing the main concepts and relations involved in the data published yearly by AMD.}
    \label{fig:ontology}
\end{figure}

The data from AMD came in the form of several \textit{csv} files and one \textit{pdf} document describing them. By considering the \textit{pdf} document, and thanks to both the contribution of the physicians of our group, and the knowledge about how the data were collected, we came up with an ontology formally describing the domain of interest. The resulting ontology has been formalised using the W3C Web Ontology Language (OWL)~\cite{W3Crec-OWL-Guide,DBLP:conf/rweb/CalvaneseGLLPRR09} and consists of all the major  relevant concepts and relations together with all the characterizing properties. Figure~\ref{fig:ontology} contains a snippet of such an ontology.

% We derived from such an ontology the schema for the database and imported all the data into this database, by means of suitable transformation and cleaning operations. All the data analysis tasks described in the next section of this paper, took their data from this remodeled and cleaned version of the data. An important contribution of the data preparation task was the definition of a domain ontology to formalize and document the knowledge of the domain of interest as well as to share such knowledge with the whole research group in an intuitive yet rigorous way. The definition of the ontology has been carried out with the contribution of domain experts, i.e. physicians specialised in diabetes treatments, and data experts, i.e. researchers that have been collaborating with AMD for several years. 

Starting from the formalization of the domain represented by the ontology, we designed a database schema trying to stay as close as possible to the conceptualization represented by the ontology. More specifically, the database schema is composed by two parts: data and metadata. The former part of the schema was meant to contain the actual data, whereas the latter contains metadata information such as the type of values that are admissible for each kind of data, the range of such values, the national level code associated with each kind of medical examination (if available), and so on.

\subsection{Data cleaning} The original data received from AMD contained several problems, inconsistencies and errors. These poor data quality problems are mainly due to the fact that AMD gathered them from real medical examinations coming from almost 300 different centres for diabetes treatment in Italy since 2005. We clarify that not all the 320 centres constituting the AMD network included their data in the latest AMD annals. Along these years, different versions of the EMR software tool have been used for collecting data, and even the same version of such a tool has been used in different ways in the various centers: this caused several semantic discrepancies in the gathered data.

A lot of cleaning actions were necessary to overcome these problems. We can distinguish the actions taken in three areas: actions to solve errors in the data, actions to solve lack and incompleteness of data, actions to preserve data privacy. For all the actions, the main criteria that have been adopted were to keep as much information as possible (or equivalently to delete the least amount of data), and to consider only the quality checks that could be done using one single table at a time, thus postponing cross-table checking to the future work. We consider the first criterion a natural approach to all data cleaning tasks, and we decided to adopt the second criterion since at this stage we only wanted to keep each table consistent with itself and with its metadata descriptions, leaving all further cleaning refinements to each specific data analytics task.

\section{Data Analysis}

Since Type 2 diabetes (T2DM) is the most prevalent form (90\% of the population with diabetes), we decided to initially focus only on subjects regularly followed at clinics and suffering from type 2 diabetes (about 600,000 subjects)
\subsection{Data-driven assessment of treatment response and disease progression}

Given the heterogeneous and multifactoral conditions characterizing type 2 diabetes worldwide, a data-driven assessment of treatment response and disease progression may help to identify potentially clinically important differences between patients. In 2018 a data-driven analysis of the Scandinavian patient registry identified five subtypes of diabetes based on clinical variables such as the age at diagnosis, BMI, HBA1c~\cite{ahlqvist2018novel}. This novel stratification method for patients was reproduced in independent cohorts in the Netherlands~\cite{gloyn2018precision} and in the UK~\cite{dennis2019disease}. We are currently using this method on the data received from AMD and we are evaluating whether it can be reproduced and whether further sub-groups emerge across the regions of Italy. 

A second objective is to study treatment response with different drug classes throughout the disease progression. For example \cite{ahlqvist2018novel} indicated that during the early stages of disease progression in insulin-resistant patients showed an increased risk of developing kidney disease, or that 
diabetic retinopathy was identified in patients with relative insulin deficiency. We are interested in applying novel techniques based on Graph Neural Networks on the prescriptions of drugs and try to identify patterns in treatment response~\cite{9122573}. Based on work done so far on the data provided by AMD, it emerges that for certain sub-groups of patients the available data are relatively sparse and often missing. Therefore it is a technical challenge to accurately detect such patterns under sparse and missing training data.

\subsection{Mining recurrent subsequences in patient's history for diabetes progression model}

The goals of this task is twofold: (i) to define a disease progression model (DPM) for diabetes, and (ii) to use this progression model to define groups of patients that have a common disease progression, and thus a common response to treatment. 
We started by looking at recurrent sequences rather then DPM, both because it is a simpler task, giving a first result on known (and unknown) diabetes behaviors, and also because such sequences would help us to have a grasp on how is the dataset we are using, whether it suffers from bias, missing data etc.
Unfortunately, recurrent sequences didn't give us reasonable results, with most of the sequences being uninformative, or unreasonable, lift $<1$, which means an improvement on the patient conditions after a certain diagnosis.
So we took a further smaller step: if we consider only 2 diagnoses in sequence (X,Y), how much is the lift? We noticed 3 things:

\begin{itemize}
    \item straightforward lift is not enough, because some subgroups (male/female,young/old) may have different lift scores, thus we need to compute lift for each group
    \item if there is a hierarchy on the organs, can we define an order? e.g. if I have a problem on peripheral arteries does this increase the risk of a major diagnosis on the hearth?
    \item A diagnosis X may be followed more often by death, which means that I would not observe the following diagnosis Y, in many dead patients. But if i observe it in all the survived patients, it should result in a very high lift, since $P(Y|X)=1$.
\end{itemize}

Subsequently, we decided to compute the lift by first splitting the population for gender and for age range (age lower or greater than a threshold). Interestingly, and unexpectedly, women seem to have a greater lift, for almost all the diagnosis pairs, compared to men. However, when you look at the percentage of diagnoses in the male and female population, men have a higher percentage, which does not contradict the lift, since the lift score is adjusted on the population size. It may be interpreted as: women are usually healthier, but once they have a first major diagnosis then the risk of having more major diagnoses increases a lot more than it does for men. Further analysis will be carried out to complete this analysis and to move towards a more generalized Diabetes progression model.

\subsection{Onset prediction of retinopathy}

Type 2 diabetes is associated with chronic microvascular complications such as diabetic nephropathy, diabetic retinopathy, and neuropathy that are the leading cause of blindness and dialysis treatment.
Microvascular complications are among the least understood. In particular, for retinopathy, there are no studied and validated prediction equations in the literature. 
The research aims to define a predictive model for the onset of retinopathy, in order to act at the level of primary prevention. 
To this aim, we consider supervised classification models based on Support Vector Machines which allow the definition of multivariate nonnlinear models.  

We select patients who have some form of retinopathy, either diagnosed or detectable by eyes examinations, and the healthy control group as the patients who are never diagnosed with the disease in their clinical history. We select 374\,283 certified non-retinopathic patients and 162\,129 certified retinopathic patients where certification consists in having a recorded diagnosis of [non]retinopathy or a [non]pathological eyes examination at least every 2 years.

A sample in the training set is constituted by a patient $p$, a visit date $t$, the patient status at time $t$ and a label $y\in \{-1,1\}$  indicating whether a diagnosis of retinopathy/non-retinopathy was made after the time $T\pm 10\% T$. 

The patient status at time $t$ is a vector defined by biographical information, values of the diagnosis and duration, prescription of diets or treatments, laboratory and instrumental analysis. 
A score accounting for the exposition of the patient to risk factors along  the clinical history up to time $t$ is also considered, so to add longitudinal cumulative information in the patient status.

\subsection{Patient clustering based on trajectory of Glycated Hemoglobin}
Glycated Hemoglobin (HbA1c) is the test of choice for diagnosing diabetes and monitoring glycemic control and raised HbA1c levels are associated with  micro and macrovascular diabetic complications. In order to understand the effects of treatments, it is crucial to investigate the trajectories of Glycated Hemoglobin over the years. Therefore, our study is looking for a trajectory-based clusterization of AMD patients. To reach this objective, we have considered 10-years trajectories of at least ten values of Glycated Hemoglobin (at least one per year), passing them to Gaussian Mixture models and Temporal K-means algorithms. Our preliminary result is a promising comparison between the clinical expertise and the identified clusters; in particular, our first attempt focused on testing various portions of the dataset with different cluster configurations to examine the differences between patients whose first HbA1c measure is closer to the diagnosis year and the others. We are currently using the above clusterization to investigate the associations of the different trajectories with several characteristics of the patient (e.g., age, drugs, BMI, microvascular and macrovascular events).

\subsection{Short-term prediction of high-risk cardiovascular events}

The objective of this task is the short-term prediction of high-risk cardiovascular events in diabetic patients, such as myocardial infarction, coronary angioplasty, and others. We leverage a dataset of over 1 million patients in the AMD dataset,  with an average of 167 recorded events (treatments, prescriptions, clinical tests and diagnoses) per patient. We model each patient's health record as a sequence of AMD event codes at irregular intervals. Since every code can be replaced by the word or phrase  describing it (e.g., \textit{fasting blood sugar, cardiovascular stress test}), we  cast the prediction task as one of next sentence prediction (NSP), a challenging problem in the domain of Natural Language Processing \cite{NSPBERT}.   To encode words and phrases associated to events, we use PubMedBERT \cite{PubMedBERT}, which is a language representation model based on BERT (Bidirectional Encoder Representations from Transformers) \cite{BERT} pre-trained on large-scale biomedical corpora. Differently from the standard NSP problem, and given the relevance of temporal information in this context, we incorporate in the  model the timestamps of recorded events\footnote{specifically, we jointly model the age of a patient and the date of occurrence of each event by associating to events their distance from the birth date.}. Furthermore, we encode drug names through their pharmacological principles, in order to highlight the similarity of drugs that are based on the same chemical substances, and we replace the numerical values of  clinical analyses with test-specific labels (\textit{normal, high,} \dots). Finally,  rather than predicting the next "words" of a sequence, we predict the probability of future occurrence of specific AMD event codes (those associate with cardiovascular risk) \textit{within} a predefined interval $\Delta$, e.g., six months.

\section{Conclusions}
In conclusion, the proposed diabetes model could be useful to understand if artificial intelligence can be at the service of health. The case of the AMD dataset poses challenges concerning the correct data interpretation, their complexity, and at the same time the need to develop different projects on the same database and perhaps define further ones in the future. We believe that spending a lot of time using AI techniques for data preparation and working side by side among AI experts and clinicians allowed us to create a unitary shared information assets that are essential to overcome problems and hopefully reach the defined clinical goals.

\section*{Acknowledgements}
The authors would like to thank the Associazione Medici Diabetologi (AMD), and the Fondazione AMD, for supporting this work. This work would have not been possible without the precious guidance and expertise of Dr. Antonio Nicolucci (CORESEARCH S.r.l.), and all patients who have been cared for over the years in the AMD centers.

\bibliographystyle{named}
\bibliography{main}

\end{document}